\pdfoutput=1
\documentclass{article}

\usepackage[dblblindworkshop, final]{neurips_2026}
\workshoptitle{1st Africa in AI Workshop}

\setcitestyle{numbers,square,comma}


\usepackage[utf8]{inputenc}
\usepackage[T1]{fontenc}
\usepackage{hyperref}
\usepackage{url}
\usepackage{booktabs}
\usepackage{tabularx}
\usepackage{placeins}
\usepackage{amsmath}
\usepackage{amssymb}
\usepackage{graphicx}
\usepackage{makecell}
\usepackage{xcolor}
\usepackage{colortbl}
\usepackage{pifont}       
\usepackage{enumitem}
\usepackage{xspace}
\usepackage{tcolorbox}
\tcbuselibrary{skins,breakable}  

\newtcolorbox{promptbox}[1]{%
  title=\textbf{#1}, enhanced, breakable,
  colback=gray!5, colframe=gray!50!black, fonttitle=\bfseries,
  fontupper=\small, boxrule=0.5pt, left=6pt, right=6pt, top=5pt, bottom=5pt,
}

\newif\ifdraft\draftfalse

\definecolor{Gray}{gray}{0.9}
\definecolor{LightCyan}{rgb}{0.88,1,1}
\definecolor{nggreen}{rgb}{0.0,0.5,0.2}

\newcommand{\cmark}{\ding{51}}
\newcommand{\xmark}{\ding{55}}

\newcommand{\naija}{NaijaFacts\xspace}
\newcommand{\yoruba}{Yor\`ub\'a\xspace}

\newcommand{\afriqa}{AfriQA\xspace}
\newcommand{\afrobench}{AfroBench\xspace}
\newcommand{\afrimmlu}{AfriMMLU\xspace}
\newcommand{\irokobench}{IrokoBench\xspace}
\newcommand{\belebele}{Belebele\xspace}
\newcommand{\simpleqa}{SimpleQA\xspace}

\newcommand{\natlas}{N-ATLaS\xspace}
\newcommand{\afrollama}{AfroLlama-V1\xspace}

\title{\naija: A Framework for Disentangling Language from Knowledge in Facts about Nigeria}

\author{%
  Zainab Afolabi  \\
  AWS FDE \\
  \texttt{zafolabi@amazon.com} \\
  \And
  Nikita Kozodoi\\
  AWS FDE \\
  \texttt{kozodoi@amazon.com} \\
  \And
  Jack Butler \\
  AWS FDE \\
  \texttt{jackbtlr@amazon.com} \\
  \And Babs Khalidson \\ AWS FDE \\ \texttt{khalbabs@amazon.co.uk}
  \And
  Sabastine Dekaa Francis\\
 Obafemi Awolowo University \\
  \texttt{sdfrancis@oauife.edu.ng}
  \And
   Christiana I. Ezeilo\\
 Obafemi Awolowo University \\
  \texttt{icezeilo@oauife.edu.ng}
  \And
  Olosunde Akinlolu Adeseye\\
  Obafemi Awolowo University \\
  \texttt{aolosunde@oauife.edu.ng}
}

\begin{document}

\maketitle

\begin{abstract}
  Multilingual benchmarks for African languages mostly translate generic questions into local languages, measuring linguistic competence but not whether models know facts about Africa. We introduce \naija, a benchmark of factual and statistical questions about Nigeria, authored by native speakers and grounded in authoritative official statistics, with every question available in English, Nigerian Pidgin, \yoruba, Igbo, and Hausa. Answering similar questions for the UK and US yields a country by language matrix that disentangles a language effect (the model knows the fact but degrades in a low-resource language) from a knowledge effect (the fact is absent in every language). Across eight frontier and Nigeria-adapted models, the knowledge effect dominates: English-query accuracy is 49--50\% for the US/UK but 20\% for Nigeria. The language effect is smaller and tracks web representation, degrading from English to \yoruba{} (9 points on Nigeria, up to 24 on US/UK questions). We find scale does not help: the strongest models reach 66--79\% on the UK and the US but none exceed 31\% on Nigerian statistics. Extending to all 54 African countries with our best performing frontier model shows the same profile continent-wide, 96\% on culture and 98\% on politics against 26\% on statistics, so the gap is not specific to Nigeria.
  \end{abstract}
\section{Introduction}
\label{sec:intro}

\begin{figure}[t]
  \centering
  \includegraphics[width=0.70\linewidth]{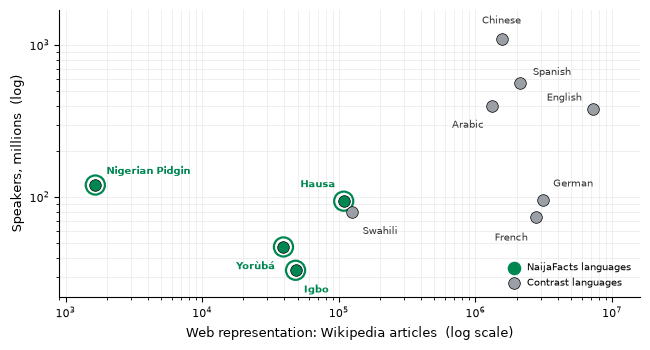}
  \caption{The \naija{} languages are widely spoken yet barely present in web
  text. Each point is a language, placed by Wikipedia
  articles~\citep{wikipedia_list2026} and speakers~\citep{ethnologue2024}, both
  on log scales. The green languages have 1--3 orders of magnitude fewer
  articles than high-resource ones with as many speakers.}
  \label{fig:representation}
\end{figure}

Large language models (LLMs) are increasingly used as sources of factual
information, yet their factual reliability is uneven and hard to
audit~\citep{OpenAI2023GPT4TR,wei2024simpleqa}. The problem compounds for
low-resource languages and for facts about under-represented regions. The
languages spoken across Nigeria illustrate the gap. As
Figure~\ref{fig:representation} shows, they are spoken by tens to hundreds of
millions of people, yet occupy a small share of the digital text on
which models are pretrained. A model can therefore fail on a Nigerian fact for
two different reasons: because it cannot process the language, or because
the fact was never well represented in its training data. Existing multilingual
benchmarks for African languages, such as \afriqa~\citep{ogundepo2023afriqa},
\afrobench~\citep{ojo2025afrobench}, and
\irokobench{}/\afrimmlu~\citep{Adelani2024IrokoBenchAN}, largely evaluate models
by translating generic questions into African languages, which does not isolate
whether a model holds grounded knowledge about Africa itself.

We argue these two failure modes should be measured separately. Consider the question, what was Nigeria's total fertility rate in the 2023? A
model can fail this in two different ways: it knows the value in
English but produces a wrong answer when asked in \yoruba{} (a language effect),
or it never encoded the value and is wrong in every language (a knowledge
effect).

We introduce \naija, a benchmark of 100 factual questions about national
demographic and health statistics, authored by native speakers of the Nigerian
languages and translated to English. Its ground truth is authoritative,
quantitative official statistics, primarily demographic and health indicators
from bodies such as the National Population Commission (NPC), the National
Bureau of Statistics (NBS), and the Nigeria Demographic and Health Survey
(NDHS). Each question is available in English, Nigerian Pidgin, \yoruba, Igbo, and Hausa, and is asked about three countries, giving 1{,}500 graded items per model. By holding the fact fixed and varying only the language of the query, \naija separates the language and knowledge effects.

A confound remains: a Nigeria-versus-English gap could reflect the resource
level of the query language instead of the model's knowledge of the country. We
therefore also vary the country a question is asked about, answering the same
questions for the United Kingdom and the United States against each country's
own official statistics. Crossing the five query languages with the three
countries gives a country $\times$ language matrix in which the language
effect is read down a column and country familiarity across a row. 

The same construction extends beyond Nigeria. Reusing the pipeline, we build
verified English question sets of comparable size for each of the 54 African
countries, drawing every answer from that country's own official statistics, so
that we can test whether the findings hold continent-wide.

We make three contributions. First, we build a benchmark that pairs
official-statistic ground truth with five-language coverage and a country
$\times$ language design, which separates the language and knowledge failure
modes. Second, we show that the knowledge effect dominates: eight models,
spanning three geopolitical origins and including two Nigeria-adapted 8B
open-weight models, hold UK and US statistics far better than Nigerian ones, and
answering in English recovers almost nothing. Third, we extend the construction
to all 54 African countries, which shows where the deficit generalises and which
facts escape it.

\section{Related Work}
\label{sec:related}

Work outside the Africa-focused research has varied the query language and the subject, but separately. Varying the query language, X-FACTR~\citep{jiang2020xfactr} and mLAMA~\citep{kassner2021mlama} probe entity-level facts with cloze prompts,
\citet{qi2023crosslingual} show that scale improves per-language accuracy but
not cross-language consistency, and \citet{aggarwal2025language} find on 10,000
country-related facts in 13 languages that a model can recall a fact in one
language and fail on it in another. Varying the subject instead,
WorldBench~\citep{worldbench2024} asks the same World Bank indicator questions
about different countries and finds sizeable geographic disparities, while
PopQA~\citep{mallen2023popqa} and Head-to-Tail~\citep{sun2024headtotail} show
recall drops sharply for entities that are less popular online.

Closest to our domain split, LocalBench~\citep{gao2025localbench} reports 56.8\% on narrative questions about US counties against under 15.5\% on numerical ones, the same asymmetry we measure across Africa (Section~\ref{sec:cross-country}). None of these crosses country with language, and most use knowledge-base triples or indicator databases as ground truth. Two further distinctions we borrow:
\citet{calderon2026recall} separate unencoded facts from encoded but
unretrieved ones, which our answer-in-English ablation tests
(Section~\ref{sec:eng-answer}), and \simpleqa~\citep{wei2024simpleqa} measures
atomic-fact correctness.

Benchmarks for African languages cover our languages but mostly not our kind of
fact. \afriqa~\citep{ogundepo2023afriqa},
\belebele~\citep{bandarkar-etal-2024-belebele},
\irokobench{}/\afrimmlu~\citep{Adelani2024IrokoBenchAN}, and
\afrobench~\citep{ojo2025afrobench} evaluate at increasing
scale on translated or generic content, testing whether a model can
operate in a language and not whether it knows facts about the region where the
language is spoken. Uhura~\citep{bayes2024uhura} moves toward knowledge with
human-translated science and truthfulness questions in six African languages,
though its content is still translated from English. Two probes are
regionally grounded: GeoMLAMA~\citep{yin2022geomlama} reports that a country's
native language is not always the best probe of knowledge about it, and
BLEnD~\citep{myung2024blend} covers everyday cultural knowledge in 13
languages including Hausa.

Afri-MCQA~\citep{tonja2026afrimcqa}, which is similar to our aim collects 7.5k
culturally grounded question-answer pairs written by native speakers across 15
African languages and, like us, sets out to separate linguistic competence from
knowledge, using external control tasks (AfriXNLI, AfriMMLU) as the language
baseline. It concludes that language understanding is the dominant limitation,
whereas we find the knowledge gap dominates. The difference is instructive: their
ground truth is everyday culture, which models largely know, whereas ours is the
quantitative official record, which they largely do not. \naija{} therefore
shares its goal and WorldBench's country axis, and adds two things neither
provides. Its ground truth is the quantitative record of a country's own
statistical office, and it varies the country asked about while holding the
question and its five language renderings fixed, with a second foreign control
that measures raw online prominence on its own. The resource skew behind the whole
gap~\citep{joshi-etal-2020-state,kudugunta2023madlad} is what
Figure~\ref{fig:representation} depicts and \naija{} measures. The nearest
neighbours each supply some of these properties but not the combination
(Table~\ref{tab:related}, Appendix~\ref{app:related}).


\section{The \naija{} Benchmark}
\label{sec:benchmark}

\naija puts 1{,}500 graded items to every model: 100 factual questions about
national demographic and health statistics, each posed in five languages and
asked about three countries, Nigeria, the United Kingdom and the United States.
The questions are authored in the Nigerian languages, not translated into them,
and every answer comes from authoritative official statistics. Because each
country is asked the same question, a difference between countries reflects the
answer, not question difficulty or phrasing. That is what separates a language
effect from a knowledge effect.

\subsection{Questions}
\label{sec:questions}
The benchmark begins from 100 questions written by native speakers and statistics
experts of \yoruba, Igbo, Hausa, and Nigerian Pidgin from Obafemi Awolowo
University. Authoring natively in each language keeps the questions fluent and
culturally natural, and avoids translationese in the low-resource languages. They
target demographic and health indicators such as population counts, fertility
rates, census methodology, and survey findings. The same team then translates
each question into English, and we review it, giving five renderings per
question. We lightly edit questions so each is self-contained and answerable for
a given country.

\subsection{Ground truth}
\label{sec:answers}
For every question we collect a gold answer for each of the three countries.
For Nigeria, the native speakers who authored the questions also provide the
answers in the Nigerian languages, grounded in the official national statistics
of the National Population Commission (NPC), National Bureau of Statistics
(NBS), and Nigeria Demographic and Health Survey (NDHS). For the United Kingdom and the United States, where these speakers are not the
domain authority, we obtain answers differently. A deep-research assistant
retrieves the relevant figure from official statistics, the Office for National
Statistics for the UK and the U.S.\ Census Bureau for the US. The four native
speakers who authored the questions then verify and correct each one. Grounding
answers in official statistics gives unambiguous, citable ground truth, and for
the UK/US answers retrieving the figure before writing it minimises fabrication.
Every item records the body, release and reference year behind its answer, and
those years span 2006 to 2025. Table~\ref{tab:sources} breaks the items down by
source.

\subsection{Coverage and item contents}
\label{sec:coverage}
Each question is instantiated for all three countries by substituting the country
it refers to, using a country-specific analog where a direct substitution would be
unnatural, and this is done in every language. The five language renderings
therefore exist for Nigeria, the UK and the US alike, so all fifteen country by
language cells hold real items rather than gaps. Each question is stored with its
text in each language, the gold answer for each country, the kind of answer it
expects (a number, a named entity, a date, or other), and the provenance of each
answer. Table~\ref{tab:datastats} summarises the dataset
(Appendix~\ref{app:datastats}) and Appendix~\ref{app:examples} gives an example
item in all five languages.

Two properties qualify what the set measures. First, the statistics cluster by
release. A model that has memorised one table can therefore answer many items at
once: 81 Nigerian items come from one survey report, and 26 US items from one
NCHS report. There are fewer independent facts here than there are items.
Second, three items ask for explanations rather than values, such as why a
post-enumeration survey follows a census. We keep them so that every item is
scored in all five languages and three countries, and we grade them more
leniently.

\subsection{Extension to 54 countries}
\label{sec:crosscountry-data}
To test whether the Nigeria results are idiosyncratic, we reuse the same recipe
to build an English-only set for every African country. Each country gets
118--125 verified items built to a fixed blueprint, which holds the domain mix
and answer-type distribution constant so that countries stay comparable. Every
answer comes from that country's own record. Statistical values are fetched
programmatically from the World Bank World Development Indicators and the DHS
Program APIs, never transcribed by a model, so a numeric gold answer cannot be
hallucinated and verifying one is simply a re-fetch. Items whose value is
missing or ambiguous at fetch time are dropped, which is why the per-country
count varies. Because instruments genuinely disagree, each indicator is pinned
to a single instrument across all countries. This yields 6{,}689 items spanning
statistics alongside smaller culture and politics subsets, which lets us ask not
just whether the deficit generalises but which kinds of fact it spares
(Section~\ref{sec:cross-country}); full provenance is in
Appendix~\ref{app:sources}.


\section{Experimental Setup}
\label{sec:setup}

We evaluate eight models (Table~\ref{tab:models}, Appendix~\ref{app:models}): six frontier models
spanning three geopolitical origins (US, EU, CN), and two Nigeria-adapted
open-weight models, \natlas{} and \afrollama, both 8B Llama-3 fine-tunes that
test whether targeted language adaptation helps. Frontier models are accessed
through AWS Bedrock; the 8B models run locally on a single 24\,GB GPU. All models share one harness and identical prompts, with a 4{,}096-token answer budget. Sampling is left at each provider's default for the hosted models, because Claude Opus 5 rejects an explicit temperature or top-$p$; the two local open-weight models are sampled at temperature 0.1.

The mainline condition is native: each question is asked in the target
language and answered in that language, as a speaker would actually use the
model; the system prompt requests only the answer, with no explanation (exact
prompts in Appendix~\ref{app:prompts}). An answer-in-English analysis
condition (Section~\ref{sec:eng-answer}) keeps the question in the target
language but requests the answer in English, testing whether failures are
expression failures. For the country analysis, each question explicitly names
the country it concerns (with a country-specific analog where a direct swap
would be unnatural) and is scored against that country's gold answer, yielding
the country $\times$ language matrix of Section~\ref{sec:results}.

Answers are free text, so we score with light normalisation and then an
LLM-as-judge (Claude Opus 5) that returns a binary verdict. It accepts
formatting, punctuation and rounding variants of the correct value, and marks
materially different figures, wrong entities, refusals and empty responses
incorrect. On the cross-country sets the judge additionally receives
per-item grading notes listing accepted variants. The full judge prompt is in
Appendix~\ref{app:judge}; unresolved verdicts are marked unclear and excluded
(9 of 16{,}000 judgments, under 0.1\%). Judged accuracy, broken down by language
and country, is the primary metric. We leave human annotators and a second judge
to future work.


\newcommand{\insertmainresults}{%
\begin{table*}[t]
\centering
\footnotesize
\begin{tabular*}{\textwidth}{@{\extracolsep{\fill}}llcccccc@{}}
\toprule
& & \multicolumn{6}{c}{Judged accuracy (\%) on \naija{} Nigeria-stats}\\
\cmidrule(lr){3-8}
Model & Size & English & Pidgin & Hausa & Igbo & \yoruba & Avg.\ (NG lang) \\
\midrule
GPT-5.6         & --   & 28 & 26 & 29 & 31 & 22 & 27.0 \\
Claude Opus 5   & --   & 28 & 26 & 21 & 25 & 24 & 24.0 \\
Grok 4.6        & --   & 29 & 23 & 25 & 25 & 19 & 23.0 \\
Kimi K2.5       & --   & 15 & 17 & 13 & \phantom{0}9 & \phantom{0}5 & 11.0 \\
Qwen3           & 235B & 18 & 16 & \phantom{0}8 & \phantom{0}8 & \phantom{0}5 & \phantom{0}9.2 \\
Mistral Large 3 & 675B & 17 & 14 & \phantom{0}7 & \phantom{0}2 & \phantom{0}2 & \phantom{0}6.2 \\
\midrule
\natlas         & 8B   & 15 & 11 & 12 & \phantom{0}6 & \phantom{0}8 & \phantom{0}9.2 \\
\afrollama      & 8B   & 11 & \phantom{0}7 & \phantom{0}6 & \phantom{0}0 & \phantom{0}5 & \phantom{0}4.5 \\
\bottomrule
\end{tabular*}
\caption{Main result, the language effect. Judged accuracy on the Nigeria
statistics set, 100 items per language, native condition. Frontier models degrade
from English in the order Pidgin $>$ Hausa $\approx$ Igbo $>$ \yoruba, tracking
web representation. The adapted 8B models (bottom block) are far weaker and
degrade in the same direction, so language adaptation does not by itself supply
Nigerian facts.}
\label{tab:main}
\end{table*}%
}

\newcommand{\insertcountrylang}{%
\begin{table}[t]
\centering
\small
\begin{tabular*}{\textwidth}{@{\extracolsep{\fill}}lccc@{}}
\toprule
& \multicolumn{3}{c}{Country asked about}\\
\cmidrule(lr){2-4}
Language asked in & Nigeria & UK & US \\
\midrule
English               & 20.1 & 49.2 & 50.2 \\
Nigerian lang.\ (avg) & 14.3 & 35.5 & 34.9 \\
\quad Pidgin          & 17.5 & 44.9 & 44.2 \\
\quad Hausa           & 15.1 & 35.2 & 35.1 \\
\quad Igbo            & 13.2 & 34.2 & 33.5 \\
\quad \yoruba         & 11.2 & 27.5 & 26.6 \\
\bottomrule
\end{tabular*}
\caption{Country by language matrix, judged accuracy (\%) averaged over all
eight models (native condition; 100 items per language per country per model).
Reading across a row isolates country familiarity at fixed language: the
Nigeria deficit is $\sim$30 points in English and persists in every language.
Reading down a column isolates the language effect at fixed country: the drop
from English to \yoruba{} is 9 points on Nigeria but 22--24 points on the
UK/US, where models actually hold the knowledge. The near-identical UK and US
columns bound the pure prominence effect at $\lesssim$1 point.}
\label{tab:countrylang}
\end{table}%
}

\section{Results}
\label{sec:results}

\subsection{Language effect vs.\ knowledge effect}
\label{sec:res-lang}
Table~\ref{tab:main} reports judged accuracy on the Nigeria statistics set by
query language in the native condition, with a per-model bar chart in
Figure~\ref{fig:bars} (Appendix~\ref{app:detailed}). The language effect is real and tracks resource level.
Averaged over the eight models, accuracy falls monotonically from English to
\yoruba{} in the order given by web representation
(Figure~\ref{fig:representation}): Pidgin, an English-derived creole, costs
under 3 points, while \yoruba, the least web-represented of the four, costs 9. The
effect is far larger for some models: Claude Opus 5, GPT-5.6, and Grok 4.6 lose
14--35\% of their English accuracy in \yoruba, whereas Qwen3, Kimi K2.5, and
Mistral Large 3 lose 67--88\%, collapsing to 2--5\% absolute.

\insertmainresults

The knowledge effect dominates the language effect.
Even in English, the best model reaches only 29\% on Nigerian statistics, while
the same three strongest models score 66--79\% on the matched US and UK sets.
Averaged over models, the English-language gap
between the US/UK and Nigeria is 30 points (Table~\ref{tab:countrylang}), more
than three times the largest language effect on the Nigeria set. Most failures on Nigerian facts are
therefore not translation failures: the fact is absent in every language.
This inverts the ordering \citet{tonja2026afrimcqa} report for African
cultural knowledge, where the language effect is the larger of the two, and the
domain is what differs. Everyday culture is well represented in web text, so
what remains to be lost there is expression, while the statistical record is
largely absent, so language has little left to subtract.

Adaptation to Nigerian languages does not supply Nigerian knowledge.
The Nigeria-adapted 8B models score 6--15\% (\natlas) and 0--11\%
(\afrollama), which is unsurprising at 30 to 100 times fewer parameters. They
also degrade across languages in the same direction and by much the same
proportion: the mean relative drop from English to \yoruba{} is 51\% for them
against 50\% for the frontier models. \natlas{} is the more stable of the two: its coefficient of variation across the
five languages is 34\% against \afrollama{}'s 68\%. The most stable models overall
are the strongest frontier ones (Claude Opus 5 10\%, GPT-5.6 13\%, Grok 4.6 15\%). Both adapted
models are weakest in Igbo, but only one of them lacks Igbo adaptation.
\afrollama{} was never trained for it, its card listing Swahili, Xhosa, Zulu,
\yoruba, Hausa and English~\citep{AfroLlamaV1}, so its 0\% there is
unsurprising. \natlas{} was: its card reports roughly 200{,}000 Igbo
instruction samples, the same budget as Hausa and \yoruba, and its developers'
own human evaluation rates its Igbo output above its
\yoruba~\citep{natlas2025}. Yet its factual accuracy is lowest in Igbo at 6\%
and twice that in Hausa at 12\%, so equal adaptation data and better-rated
fluency did not translate into better factual recall. Fine-tuning
an 8B model on Nigerian languages therefore leaves both gaps open at this
scale: it neither supplies the factual record nor buys robustness to the query
language. Closing the factual gap appears to need frontier scale or explicit
injection of African knowledge during training.

The deficit does not depend on where a model was built. Origin does separate the
models on Nigerian statistics: asked in English, the US-built systems average
28.3\% against 17.0\% for the European and 16.5\% for the Chinese ones. But that
ordering is the same on the UK and US sets, so it tracks capability rather than
whose facts a model holds. The test that would show provenance is a home-country
advantage, and there is none: the three US-built models score $+1$, $-8$ and
$+7$ points on the US set relative to the UK set. No model of any origin exceeds
31\% on Nigerian statistics. The converse test is unavailable, since the benchmark
has no French or Chinese question set.

\subsection{Country $\times$ language}
\label{sec:res-country}
\begin{figure}[t]
  \centering
  \includegraphics[width=0.60\linewidth]{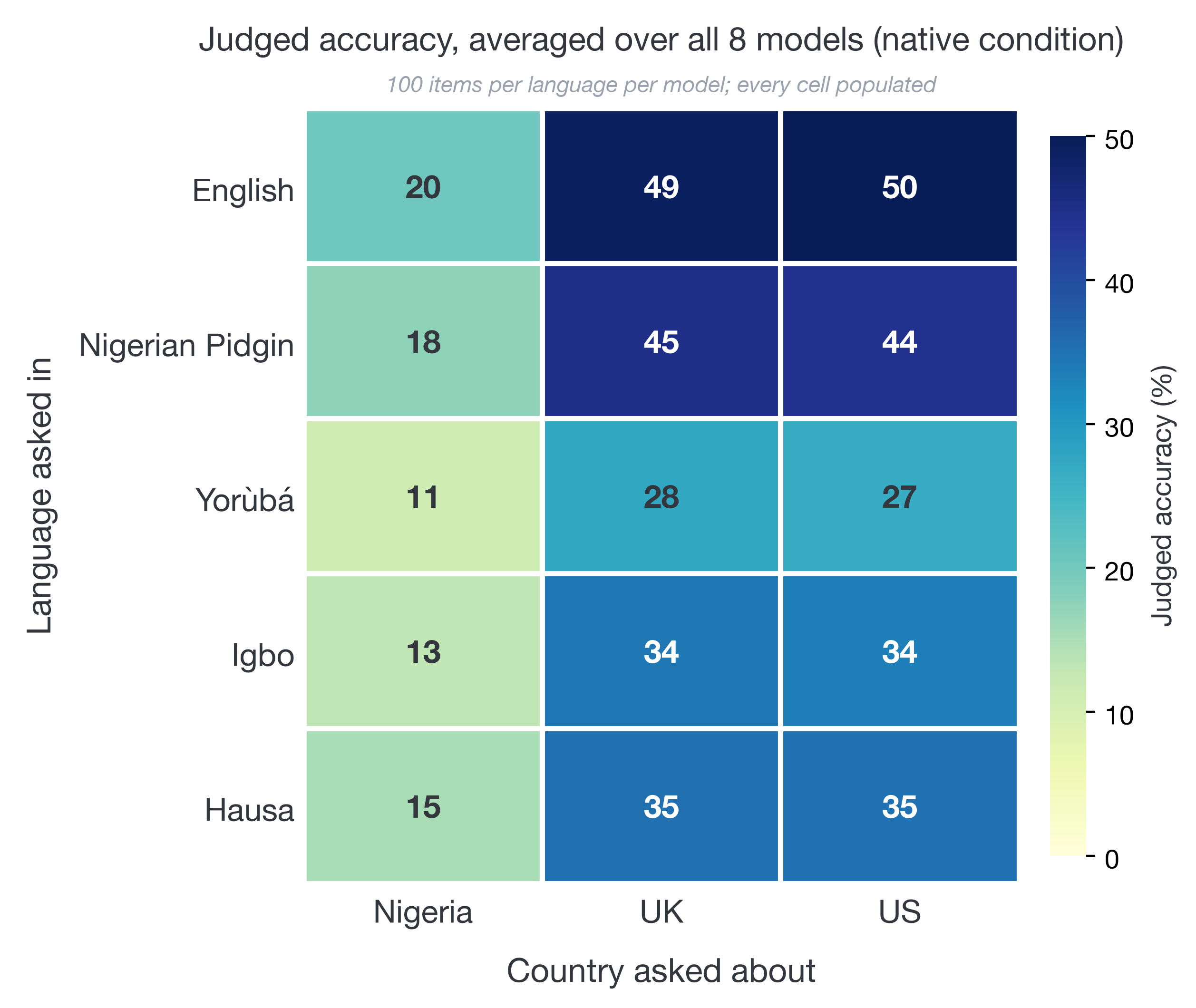}
  \caption{Country $\times$ language judged accuracy, averaged over all eight
  models (native condition). Columns (country asked about) separate far more
  than rows (language asked in): the knowledge effect dominates. The worst cell
  is \yoruba$\times$Nigeria, where both effects compound.}
  \label{fig:heatmap}
\end{figure}

Figure~\ref{fig:heatmap} shows accuracy as a function of the country asked about
and the language asked in, with the underlying numbers in
Table~\ref{tab:countrylang} (Appendix~\ref{app:detailed}).

The expected ordering of the four conditions holds, model-averaged:
$A(\text{en}, \text{US/UK}) \geq A(\text{NG lang}, \text{US/UK}) \geq
A(\text{en}, \text{Nigeria}) \geq A(\text{NG lang}, \text{Nigeria})$. Since
\yoruba{} and Nigeria co-occur throughout the training text, asking about Nigeria
in \yoruba{} might have been the pairing models handle best. It is their weakest
instead, at 11.2\%.

Read across rows, at fixed language, the Nigeria deficit is present in all five
languages but its size varies with the language: 30 points in English and 27 in
Pidgin, narrowing to 20--21 in Hausa and Igbo and 16 in \yoruba. Read down
columns, at fixed country, the English-to-\yoruba{} drop is 9 points on Nigeria
but 22--24 on the UK/US. Both directions point the same way. Language costs most
where the model has knowledge to lose, and the country deficit shrinks in the
languages where models are already wrong in English.

Pure country prominence, by contrast, is near zero. The US-versus-UK gap at fixed
language, our control for raw online prominence, is at most 1 point in any
language (50.2 vs 49.2 in English, 26.6 vs 27.5 in \yoruba). The US is far more
present online than the UK, yet the models score them the same, so online
prominence cannot be what produces the 30-point Nigeria deficit. What separates
Nigeria from these two is how well the models know the country.

\section{Analysis and Ablations}
\label{sec:analysis}

\subsection{Answering in English does not recover accuracy}
\label{sec:eng-answer}
If a model knows a fact but cannot express it in \yoruba, letting it answer in
English should recover the accuracy. We re-ran the full Nigeria set with
questions still asked in each Nigerian language but answers requested in
English. The effect is negligible: averaged over the eight models, the change
from the native condition is $+0.5$ points for Pidgin, $-0.9$ for Hausa,
$-0.2$ for Igbo, and $+1.9$ for \yoruba. Expression is not the bottleneck. A
model that fails on a Nigerian fact asked in \yoruba{} does not know the fact,
and is not merely unable to say it in \yoruba. If translation were the problem,
removing it would have recovered the accuracy. Figure~\ref{fig:recovery} compares
the two conditions per model.

\subsection{The language effect tracks web representation}
\label{sec:resource-corr}
\begin{figure}[t]
  \centering
  \includegraphics[width=0.68\linewidth]{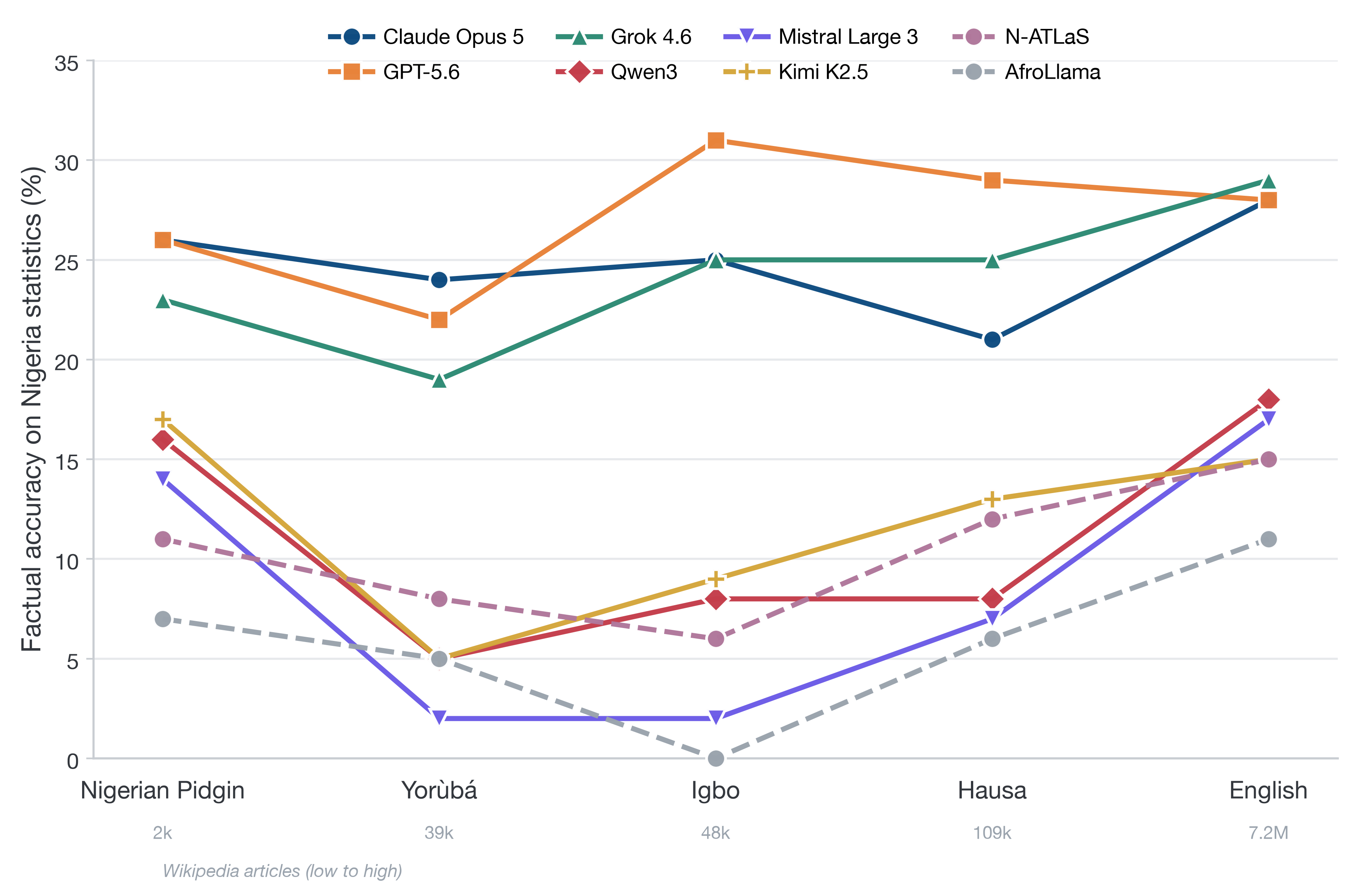}
  \caption{Model-averaged judged accuracy per query language against the
  language's web representation (Wikipedia articles, log scale), for each
  country set. Accuracy rises with representation; Pidgin sits far above the
  trend for its article count, consistent with its legibility through
  English.}
  \label{fig:acc-vs-rep}
\end{figure}

Figure~\ref{fig:acc-vs-rep} plots per-language accuracy against each
language's Wikipedia article count (a proxy for web text availability;
Figure~\ref{fig:representation}). Averaged over models, the ordering
English $>$ Pidgin $>$ Hausa $>$ Igbo $>$ \yoruba{} is monotone on all three
country sets. Two details sharpen the picture. First, Pidgin behaves unlike the
other three. It has the fewest Wikipedia articles, about 1.6k, yet costs the
least, because an English-derived creole is largely legible through English
itself. Article counts understate its effective representation. Second, the slope depends on how much knowledge there is to
lose: the English-to-\yoruba{} drop is 22--24 points on the UK/US sets but
only 9 points on Nigeria, where English accuracy is already low. Language
degrades the retrieval of knowledge the model has; it cannot subtract
knowledge the model lacks. Figure~\ref{fig:langcost} in Appendix~\ref{app:detailed} breaks this down per
model. Every model holds substantial UK/US knowledge in English, but how much
survives a Nigerian query language varies sharply with capability: Claude Opus 5
and GPT-5.6 keep nearly all of it, Qwen3 and Mistral Large 3 keep about a third.

\subsection{The deficit extends across the continent}
\label{sec:cross-country}
We evaluated Claude Opus 5 on the 54 per-country sets of
Section~\ref{sec:crosscountry-data}. Figure~\ref{fig:africa-map} maps the
result, and Table~\ref{tab:country-accuracy} in Appendix~\ref{app:detailed}
lists every per-country accuracy.

The domain split is large. The model scores 95.8\% on culture (689 items) and
97.7\% on politics (642), but 26.0\% on statistics (5{,}358). Frontier models
hold the salient, narrative facts about African
countries, meaning capitals, leaders, and landmarks, while the quantitative
record of fertility, mortality, and economic indicators from national
statistical offices and DHS surveys is almost entirely absent. A benchmark built
from salient facts would conclude that models know Africa well, and grounded
statistics show they do not.

\begin{figure}[t]
  \centering
  \includegraphics[width=0.80\linewidth]{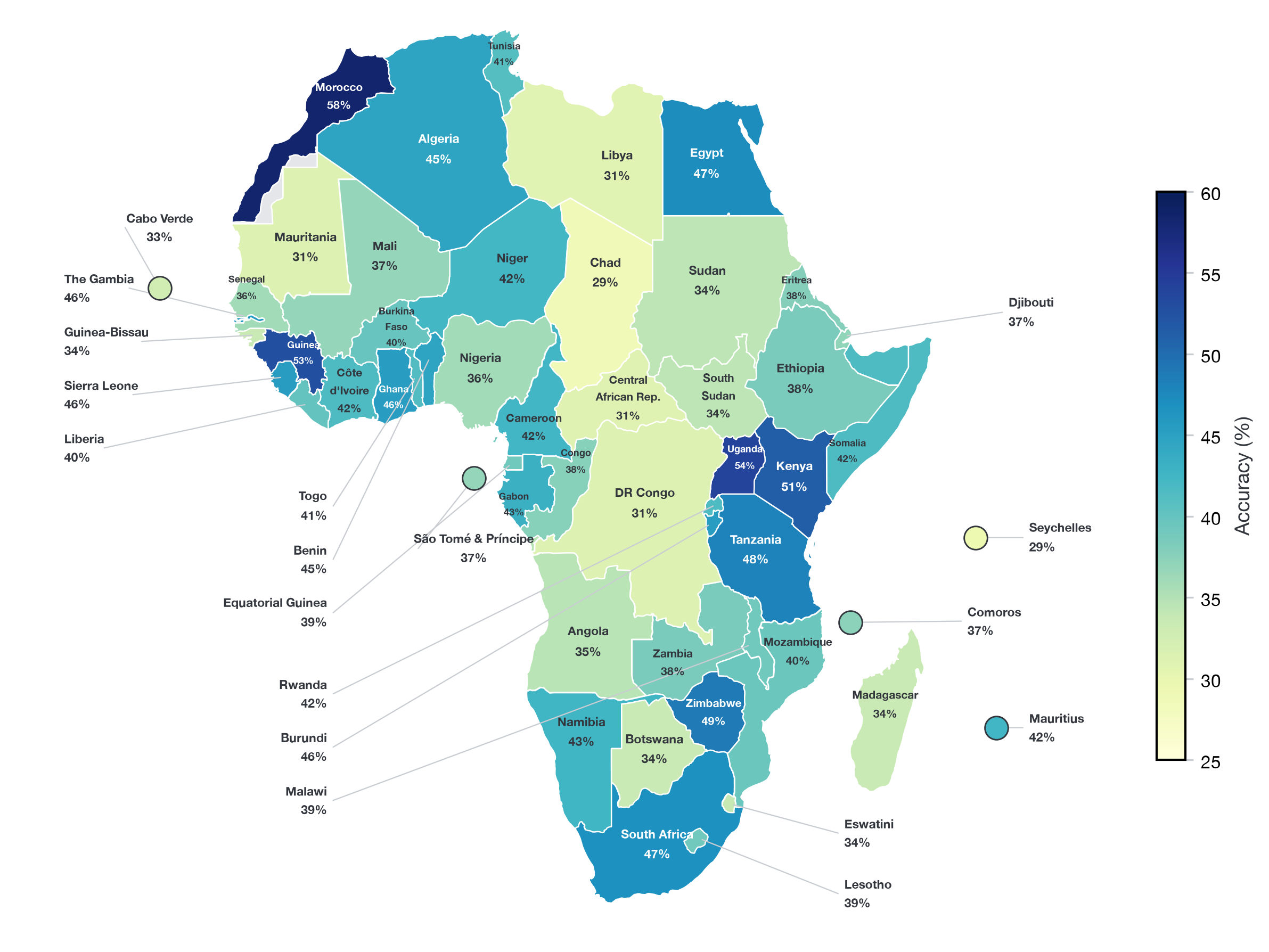}
  \caption{Judged accuracy of Claude Opus 5 on country-grounded English
  questions across all 54 African countries (118--125 items each), spanning
  28.8\% (Chad) to 58.4\% (Morocco), mean 40.1\%.}
  \label{fig:africa-map}
\end{figure}

Aggregate accuracy is 40.1\% (2{,}681/6{,}689), but statistics is the bulk of
the set, so that number is close to the statistics figure pulled up by two
small, easy domains. Per country it ranges from 28.8\% (Chad) to 58.4\%
(Morocco), and Nigeria, Africa's most populous country, sits below the middle at
36.0\%. Nigeria scores higher here than the 28\% of Section~\ref{sec:results}
only because this set mixes in the two easy domains. The deficit is therefore
continental and not Nigerian.

The same split appears inside the Nigeria set. Eleven items ask about the
statistical system rather than a statistic, such as what TFR abbreviates, and
they score 59.8\% against 8.6\% on the items drawn from the 2024 NDHS. Those
eleven are also the only Nigerian items with non-numeric answers, so salience
and answer type cannot be separated here.

A stale training cut-off does not explain the deficit. Items about the 2006
census score 9.2\%, no better than the 2024 items, so eighteen-year-old Nigerian
statistics are absorbed no better than last year's. Age is not what makes these
facts hard.

A similar asymmetry has been measured outside Africa.
LocalBench~\citep{gao2025localbench} reports 56.8\% on narrative questions about
US counties against under 15.5\% on numerical ones, so part of what we see is a
general difficulty with quantitative facts and not specific to Africa. It cannot
be the whole story, though, because the UK and US sets beat every African country
on comparable facts. Both effects are present, and the geographic one survives
after the numeric one is accounted for.

We think the explanation is the form in which these statistics exist, not whether
they exist at all. Every figure we grade is public, but it lives inside a PDF
report table, and pretraining corpora are dominated by running prose, so a number
can be openly available and still barely present in what a model absorbs. That
would explain why contamination looks unlikely even though every source is open,
and it predicts which fix works: retrieval over authoritative tables rather than
more pretraining.

\section{Conclusion}
\label{sec:conclusion}

\naija{} is a multilingual benchmark that grounds factual questions about
Nigeria in authoritative official statistics and separates two failure modes
that translation-based benchmarks conflate: degradation in a low-resource
language, and absence of the underlying fact. Asking the same questions about
the UK and the US separates country familiarity from language resource level.

The knowledge effect dominates. Asked in English, models average 49--50\% on US
and UK official statistics but only 20\% on the matched Nigerian statistics, and
answering in English recovers essentially nothing, so most failures on Nigerian
facts reflect missing knowledge and not translation. Scale does not rescue it
either: the strongest models reach 66--79\% on the UK and US sets while no model
exceeds 31\% on Nigerian statistics. The language effect is real but smaller and
tracks web representation, falling from English through Pidgin, Hausa, and Igbo
to \yoruba, and it costs most where models have knowledge to lose. Across all 54
African countries the profile is the same, with 96\% accuracy on culture and
98\% on politics against 26\% on official statistics.

Serving African users therefore needs authoritative African data, the
statistical record in particular, curated into training corpora, not only
models adapted to African languages. We release the data, harness, and per-item
outputs so the benchmark can be reproduced and extended to further countries.

Our study has a few limitations that future work should address.
\label{sec:limitations}
The Nigeria subset is 100 items per language drawn largely from a few source
tables, so it probes grounded statistical knowledge and not broad knowledge, and
the 54-country extension is English-only on a single model. Official statistics
are also public, so contamination cannot be excluded, though the low accuracy on
statistics suggests these tables are poorly absorbed in practice. The next steps
follow: extending the multilingual design beyond Nigeria, and testing whether
injecting statistical tables during training moves the number that language
adaptation leaves fixed.

\FloatBarrier

\bibliographystyle{plainnat}
\bibliography{references}

\appendix
\section{Prompt templates}
\label{app:prompts}

Every model sees the same two prompts in every condition. Placeholders in braces
are substituted per item: \texttt{\{language\}} names the question's language
(English, Hausa, Yoruba, Igbo, or Nigerian Pidgin) and \texttt{\{question\}}
carries the question text in that language.

\begin{promptbox}{System prompt (all conditions)}
You are a factual question-answering assistant. IMPORTANT: Reply with ONLY the
answer itself, no explanations, no preamble, no `The answer is...', no additional
context.

For numbers: give just the number (e.g., `200 million' or `45.2\%').

For names/entities: give just the name (e.g., `Abuja' or `National Population
Commission').
\end{promptbox}

\begin{promptbox}{User prompt (native condition)}
Answer this question in \{language\}:

\{question\}

Output format: Give only the answer, nothing else.
\end{promptbox}

The answer-in-English condition (Section~\ref{sec:eng-answer}) changes one word
of the user prompt, so that the first line reads ``Answer this question in
English:'' while the question itself stays in the target language. Everything
else, including the system prompt, is unchanged.

\section{LLM-as-judge prompt}
\label{app:judge}

The judge (Claude Opus 5) receives the system prompt below and returns a
single-token verdict. The user turn supplies the question, the gold answer, and
the model's answer.

\begin{promptbox}{Judge system prompt}
You are an evaluator assessing whether a model's answer is factually correct
compared to the ground truth.

Your task: Determine if the model's answer contains the same core factual
information as the expected answer.

IMPORTANT -- Mark as CORRECT if:
\begin{itemize}[nosep, leftmargin=1.2em, label=--]
  \item The core fact matches, even with different wording
  \item Numbers are the same or very close (89\% = 88.9\%, 59597542 = 59,597,300)
  \item Minor additions/omissions of words like ``The'', ``a'', ``an'', articles
  \item The meaning is the same even if the actual words are different
  \item Missing or extra punctuation (periods, commas)
  \item Missing units when the number is correct (``1.41'' = ``1.41 children per woman'')
  \item Abbreviations match full names (``ONS'' = ``Office for National Statistics'')
  \item Extra context is provided but core answer is correct
\end{itemize}

Mark as INCORRECT ONLY if:
\begin{itemize}[nosep, leftmargin=1.2em, label=--]
  \item The factual content is genuinely wrong
  \item The answer is empty or refuses to answer
  \item A completely different fact is stated
\end{itemize}

Examples of CORRECT matches:
\begin{itemize}[nosep, leftmargin=1.2em, label=--]
  \item ``Office for National Statistics'' vs ``The Office for National Statistics.''
  \item ``Every 10 years'' vs ``Every 10 years.''
  \item ``89\%'' vs ``88.9\%''
  \item ``56,489,800'' vs ``56,489,800 people.''
  \item ``Census Act 1920'' vs ``The Census Act 1920.''
  \item ``London'' vs ``London.''
\end{itemize}

Respond with ONLY: CORRECT or INCORRECT
\end{promptbox}

For the native condition the judge compares in the answer's language, and for the
answer-in-English condition it compares in English. On the 54-country sets the
user turn carries one extra block, the item's grading notes, which name the answer
variants to accept and the plausible-sounding answers the sources positively
contradict. Those notes override the defaults above, so an answer the notes rule
out is marked incorrect even when it looks reasonable
(Section~\ref{sec:setup}).

\section{Example items}
\label{app:examples}

Table~\ref{tab:example-item} shows one \naija{} item in all five languages.
Every item carries the same gold answer across languages, so any accuracy
difference between languages on the same item is attributable to the query
language alone. Diacritics are reproduced from the source data.
Table~\ref{tab:example-numeric} shows a numeric item in the same format,
together with its gold answer in each language and how many models answered it
correctly, which reproduces the paper's main language ordering within a
single item.

\begin{table}[b]
\centering
\small
\begin{tabularx}{\textwidth}{lX}
\toprule
Language & Question (gold answer: National Population Commission) \\
\midrule
English & What does NPC stand for in Nigeria's population statistics system? \\
Pidgin  & Wetin be di NPC stand for for Naija population statistics? \\
\yoruba & Kini NPC duro fun ninu eto i\d{s}iro awon eniyan Naijiria? \\
Igbo    & G\d{i}n\d{i} ka NPC p\d{u}tara n'usoro \d{o}n\d{u} \d{o}g\d{u}g\d{u} nd\d{i} mmad\d{u} nke Na\d{i}jir\d{i}a? \\
Hausa   & Menene ma'anar NPC a tsarin kididdigar yawan jama'a na Najeriya? \\
\bottomrule
\end{tabularx}
\caption{An example \naija{} item across the five languages. Most items have
numeric gold answers drawn from official statistics (e.g.\ ``How many
population classifications were used in the 2006 Nigeria census?'' with gold
answer ``Eight''); entity items such as this one are included for coverage of
the statistical system itself.}
\label{tab:example-item}
\end{table}

\begin{table}[t]
\centering
\small
\begin{tabularx}{\textwidth}{lXXc}
\toprule
Language & Question & Gold answer & Models correct \\
\midrule
English    & What age range was used for all Nigerian women interviewed in the 2024 NDHS? & 15 to 49 years. & 8/8 \\
Pidgin     & Wetin be di age range wey dem use for all Naija wumen wey dem interview for 2024 NDHS? & 15 to 49 years. & 7/8 \\
Hausa      & Wane adadin shekarun matan Nageriya aka yi wa tambayoyi a NDHS 2024? & Shekaru 15 zuwa 49. & 6/8 \\
Igbo       & Kedu oke af\d{o} e jiri maka \d{u}m\d{u} nwany\d{i} Naijira niile nd\d{i} a gbara aj\d{u}j\d{u} \d{o}n\d{u} na NDHS 2024? & Af\d{o} 15 ruo 49. & 5/8 \\
\yoruba    & Kini iw\d{o}n \d{o}j\d{o}-ori awon obinrin ni Naijiria ti won ba s\d{o}r\d{o} ninu NDHS 2024? & \d{O}dun 15 si 49. & 4/8 \\
\bottomrule
\end{tabularx}
\caption{A numeric item in all five languages, with the gold answer as it
appears in each language and the number of the eight evaluated models judged
correct on it. The fact is identical in every row, and the answer is a short
figure that leaves little room for expression difficulty, yet the count falls
monotonically in exactly the order reported in
Section~\ref{sec:res-lang}: English, Pidgin, Hausa, Igbo, \yoruba. Note that
the gold answer is itself translated, so numerals stay Arabic while the
surrounding words change (\d{O}dun, Af\d{o}, Shekaru); the judge accepts any of
these forms.}
\label{tab:example-numeric}
\end{table}

\section{Dataset summary}
\label{app:datastats}

Table~\ref{tab:datastats} summarises \naija{} and its cross-country extension.
The two halves are matched in different ways, which is worth keeping in mind when
reading any comparison between them. The Nigeria, UK and US sets ask the same 100
questions of all three countries, so a difference between countries is
attributable to the answer alone. The 54-country extension cannot work that way,
because a question about a national census has no counterpart in a country that
has never run one, so it instead holds the domain mix and answer-type
distribution constant and draws each country's answers from its own sources.

\newcommand{\insertdatastats}{%
\begin{table}[t]
\centering
\small
\begin{tabularx}{\textwidth}{lX}
\toprule
Component & Value \\
\midrule
Questions & 100, authored by native speakers \\
Question languages & \yoruba, Igbo, Hausa, Nigerian Pidgin (native); English (translated) \\
Answer countries & Nigeria (NPC/NBS/NDHS), UK (ONS), US (Census Bureau/NCHS) \\
Answer collection & Nigeria: native speakers, in-language; UK/US: deep-research, verified \\
Evaluation cells & 100 $\times$ 5 languages $\times$ 3 countries $=$ 1{,}500 per model \\
Cross-country extension & 54 African countries, 118--125 English items each (6{,}689 total) \\
\bottomrule
\end{tabularx}
\caption{\naija{} at a glance. The same questions are answered
for three countries, so the country comparison is matched by construction; the
54-country extension instead matches a fixed blueprint (domain mix,
answer-type distribution) with each country's facts drawn from its own
official sources.}
\label{tab:datastats}
\end{table}%
}

\insertdatastats

\section{Models evaluated}
\label{app:models}

Table~\ref{tab:models} lists the eight models, their origins and their
identifiers. The frontier set spans three geopolitical origins so that we can ask
whether a model's origin affects whose facts it holds
(Section~\ref{sec:res-lang}), and the two Nigeria-adapted models are both
Llama-3-8B fine-tunes, which lets us separate the effect of language adaptation
from the effect of scale. Parameter counts for the proprietary models are
undisclosed. Table~\ref{tab:countrylang} then gives the model-averaged country by
language matrix discussed in Section~\ref{sec:res-country}.

\newcommand{\insertmodels}{%
\begin{table}[t]
\centering
\small
\begin{tabular*}{\textwidth}{@{\extracolsep{\fill}}lllc@{}}
\toprule
Model & Origin & Identifier & Size \\
\midrule
\multicolumn{4}{l}{Frontier (proprietary, via AWS Bedrock)}\\
Claude Opus 5   & US & \texttt{anthropic.claude-opus-5}                & -- \\
GPT-5.6         & US & \texttt{openai.gpt-5.6-sol}                     & -- \\
Grok 4.6        & US & \texttt{xai.grok-4.6}                           & -- \\
Mistral Large 3 & EU & \texttt{mistral.mistral-large-3-675b-instruct}  & 675B \\
Qwen3           & CN & \texttt{qwen.qwen3-235b-a22b-2507}              & 235B \\
Kimi K2.5       & CN & \texttt{moonshotai.kimi-k2.5}                   & -- \\
\midrule
\multicolumn{4}{l}{Nigeria-adapted (open weight, Llama-3-8B fine-tunes, local)}\\
\natlas      & NG & \texttt{NCAIR1/N-ATLaS}          & 8B \\
\afrollama   & NG & \texttt{Jacaranda/AfroLlama\_V1} & 8B \\
\bottomrule
\end{tabular*}
\caption{Models evaluated. The frontier set spans three geopolitical origins
(US, EU, CN), enabling the question of whether a model's origin affects whose
facts it encodes; the Nigeria-adapted models are 8B Llama-3 fine-tunes
targeting Nigerian languages. Proprietary parameter counts are undisclosed.
Cross-region inference-profile prefixes are omitted from the identifiers.}
\label{tab:models}
\end{table}%
}

\insertmodels

\insertcountrylang

\section{Benchmark comparison}
\label{app:related}

Table~\ref{tab:related} compares \naija{} against the nearest related
benchmarks along the three properties discussed in
Section~\ref{sec:related}: grounded statistical answers, Nigerian-language
coverage, and a matched country $\times$ language design. No prior benchmark
combines all three, which is the gap \naija{} fills. The nearest neighbours each
supply one or two: WorldBench varies the country but uses indicator databases and
English only, Afri-MCQA authors natively in African languages but grounds its
answers in everyday culture, and LocalBench grounds answers in official
statistics but for US counties in English.

\newcommand{\insertrelatedworks}{%
\begin{table*}[t]
\centering
\footnotesize
\newcolumntype{Y}{>{\raggedright\arraybackslash}X}
\begin{tabularx}{\textwidth}{@{}lccYcY@{}}
\toprule
Benchmark & \# Afr.\ lang. & \makecell{Facts about\\Africa?} & \makecell{Ground-truth\\source} & \makecell{Country $\times$\\lang.?} & Dominant task(s) \\
\midrule
\afriqa~\cite{ogundepo2023afriqa}            & 10 & partial & Wikipedia / web & \xmark & Cross-lingual QA \\
\belebele~\cite{bandarkar-etal-2024-belebele} & 29 & \xmark & translated passages & \xmark & Reading comprehension \\
\irokobench~\cite{Adelani2024IrokoBenchAN}    & 17 & \xmark & translated (MMLU/\allowbreak MGSM/\allowbreak NLI) & \xmark & MMLU, Math, NLI \\
\afrobench~\cite{ojo2025afrobench}            & 64 & \xmark & aggregated / translated & \xmark & 15 NLP tasks \\
Uhura~\cite{bayes2024uhura}                  & 6 & \xmark & translated ARC / TruthfulQA & \xmark & Science QA, truthfulness \\
Afri-MCQA~\cite{tonja2026afrimcqa}           & 15 & \cmark~(culture) & native-authored culture QA & \xmark & Multimodal cultural QA \\
GeoMLAMA~\cite{yin2022geomlama}              & 1 (Swahili) & partial & templated commonsense & partial & Commonsense probing \\
BLEnD~\cite{myung2024blend}                  & \makecell{1 (Hausa)} & partial & hand-crafted culture QA & \xmark & Everyday knowledge \\
WorldBench~\cite{worldbench2024}             & \xmark & partial & World Bank indicators & \xmark~(country only) & Country factual recall \\
LocalBench~\cite{gao2025localbench}          & \xmark & \xmark & US Census, news, Reddit & \xmark~(county only) & Local knowledge, reasoning \\
\midrule
\rowcolor{Gray}
\naija~(Ours) & 5 & \cmark~(Nigeria) & official statistics (NPC, NBS, NDHS) & \cmark~(NG/UK/US) & Grounded factual / stats QA \\
\bottomrule
\end{tabularx}
\caption{\naija{} in context. Unlike prior work, \naija{} grounds every item in
authoritative national statistics about Nigeria and crosses the country asked
about with the language asked in. GeoMLAMA and BLEnD are the closest
geo/cultural probes but use commonsense or everyday-culture ground truth and do
not provide a matched country by language control. Afri-MCQA shares our aim of
separating language from knowledge but grounds answers in culture; LocalBench
grounds answers in official statistics but for US counties in English only.
\belebele{} reports 122 language variants in total but no African subtotal; the
29 given here is our own count of its variants indigenous to or principally
spoken in Africa, which includes Afrikaans and the Moroccan and Egyptian Arabic
variants.}
\label{tab:related}
\end{table*}%
}

\insertrelatedworks

\section{Sources and provenance}
\label{app:sources}

\begin{table}[b]
\centering
\small
\begin{tabularx}{\textwidth}{lXr}
\toprule
Country & Source body & Items \\
\midrule
Nigeria & Nigeria Demographic and Health Survey 2024~\citep{ndhs2024} & 81 \\
        & NPC census documentation, 2006 and 2023~\citep{npc}         &  8 \\
        & NBS releases carrying no edition year~\citep{nbs}           & 11 \\
\midrule
UK      & Office for National Statistics~\citep{ons}                  & 56 \\
        & UK government departments                                   & 11 \\
        & National Records of Scotland and Scotland's Census~\citep{ons} &  9 \\
        & WHO/UNICEF Joint Monitoring Programme~\citep{jmp}           &  8 \\
        & UK legislation                                              &  5 \\
        & other official sources                                      & 11 \\
\midrule
US      & National Center for Health Statistics~\citep{nchs}          & 42 \\
        & U.S.\ Census Bureau~\citep{uscensus}                        & 38 \\
        & WHO/UNICEF Joint Monitoring Programme~\citep{jmp}           &  9 \\
        & U.S.\ Environmental Protection Agency                       &  4 \\
        & other official sources                                      &  7 \\
\bottomrule
\end{tabularx}
\caption{Where the gold answers come from, by country. Reference years span 2006
to 2025 and are recorded in the bibliography entry for each body. Every UK and US
item additionally stores the URL of the specific release it was taken from, and
those URLs are distributed with the benchmark; the Nigerian items name their
census or survey edition in the question text.}
\label{tab:sources}
\end{table}

Table~\ref{tab:sources} gives the per-country breakdown. All sources are public
official statistics. The 54-country extension instead draws on the World Bank
World Development Indicators and the DHS Program APIs
(Section~\ref{sec:crosscountry-data}); one example of the instrument
disagreement that forces each indicator to a single source is that WDI and
national DHS reports differ on total fertility rates.

\section{Per-language and per-model results}
\label{app:detailed}

Three tables carry the full numbers behind Sections~\ref{sec:results}
and~\ref{sec:analysis}: Table~\ref{tab:main} the per-model by per-language
breakdown on the Nigeria set, Table~\ref{tab:countrylang} the model-averaged
country by language matrix, and Table~\ref{tab:country-accuracy} the per-country
accuracies for all 54 African countries.

\begin{figure}[t]
  \centering
  \includegraphics[width=\linewidth]{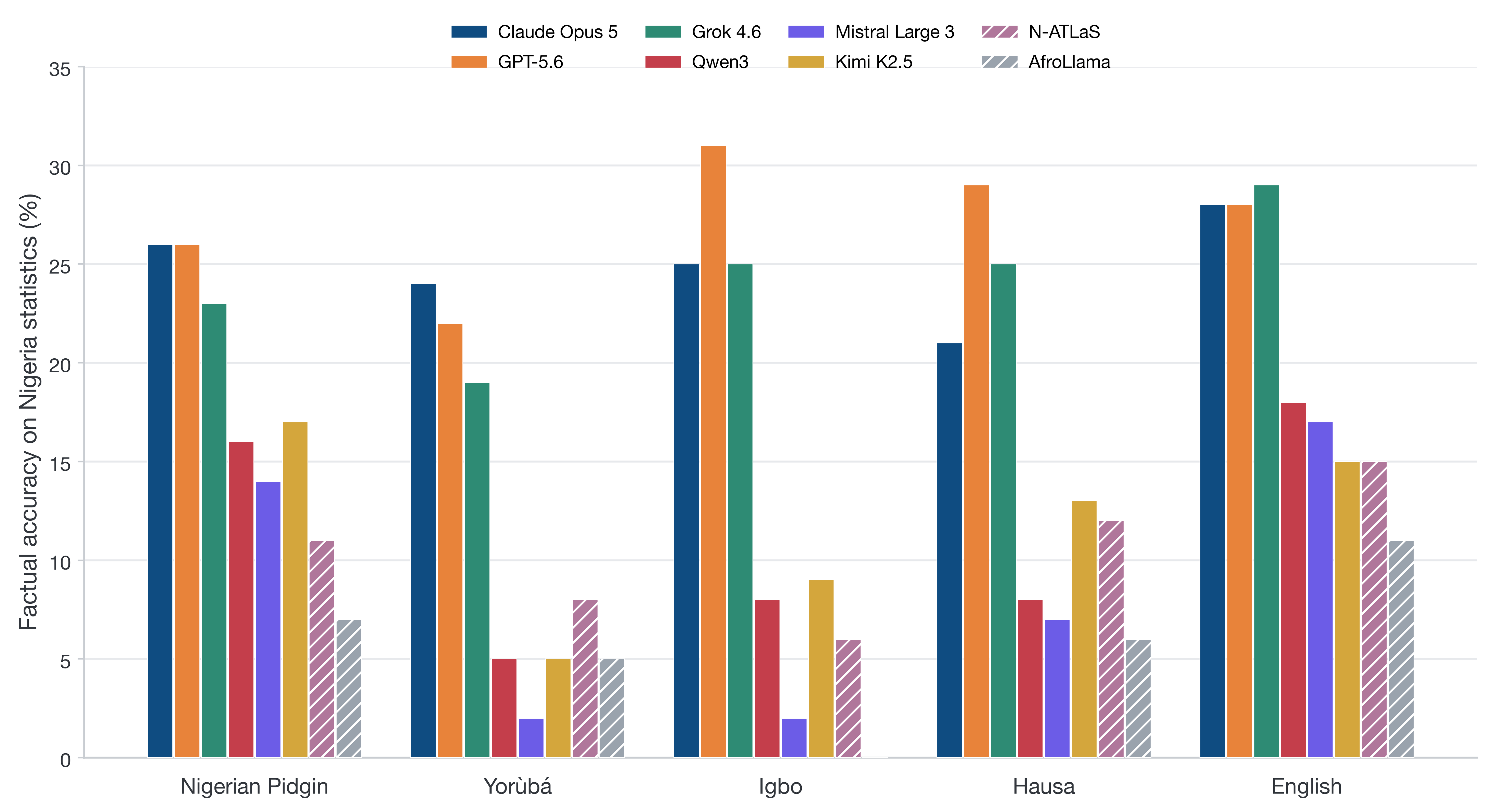}
  \caption{Judged accuracy on the Nigeria statistics set by query language
  (native condition), one bar per model within each language cluster;
  languages ordered by web representation and Nigeria-adapted models hatched.
  Frontier models degrade as representation falls, while the adapted 8B
  models stay flat across languages at a lower level.}
  \label{fig:bars}
\end{figure}

\begin{figure}[b]
  \centering
  \includegraphics[width=\linewidth]{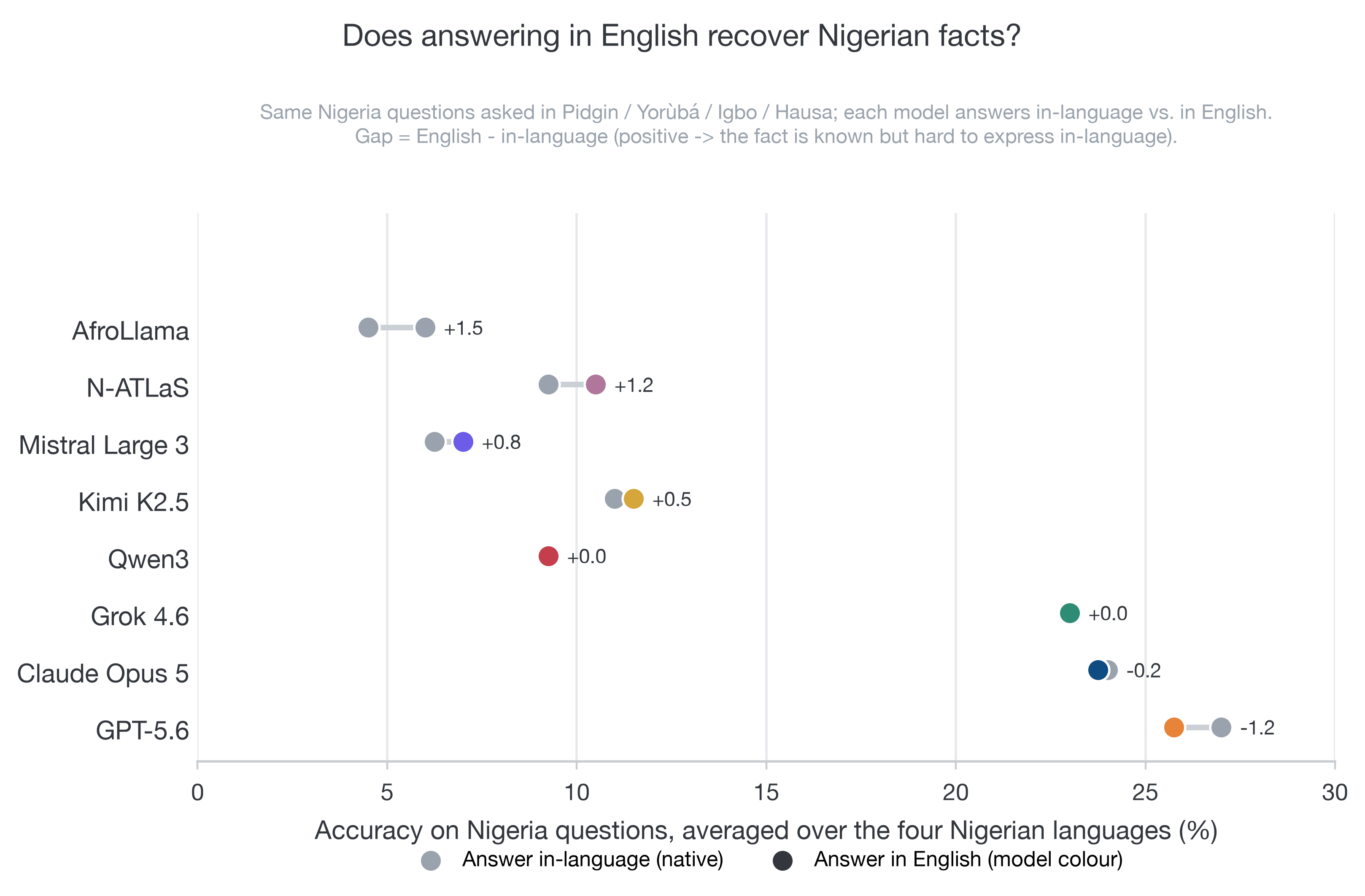}
  \caption{Answering in English does not recover Nigerian facts. For each
  model, accuracy on Nigeria questions averaged over the four Nigerian
  languages when answering in-language (grey) versus in English (colour). The
  gaps are within $\pm 2$ points for every model: failures reflect missing
  knowledge, not an inability to express the answer in-language.}
  \label{fig:recovery}
\end{figure}

\begin{figure}[t]
  \centering
  \includegraphics[width=\linewidth]{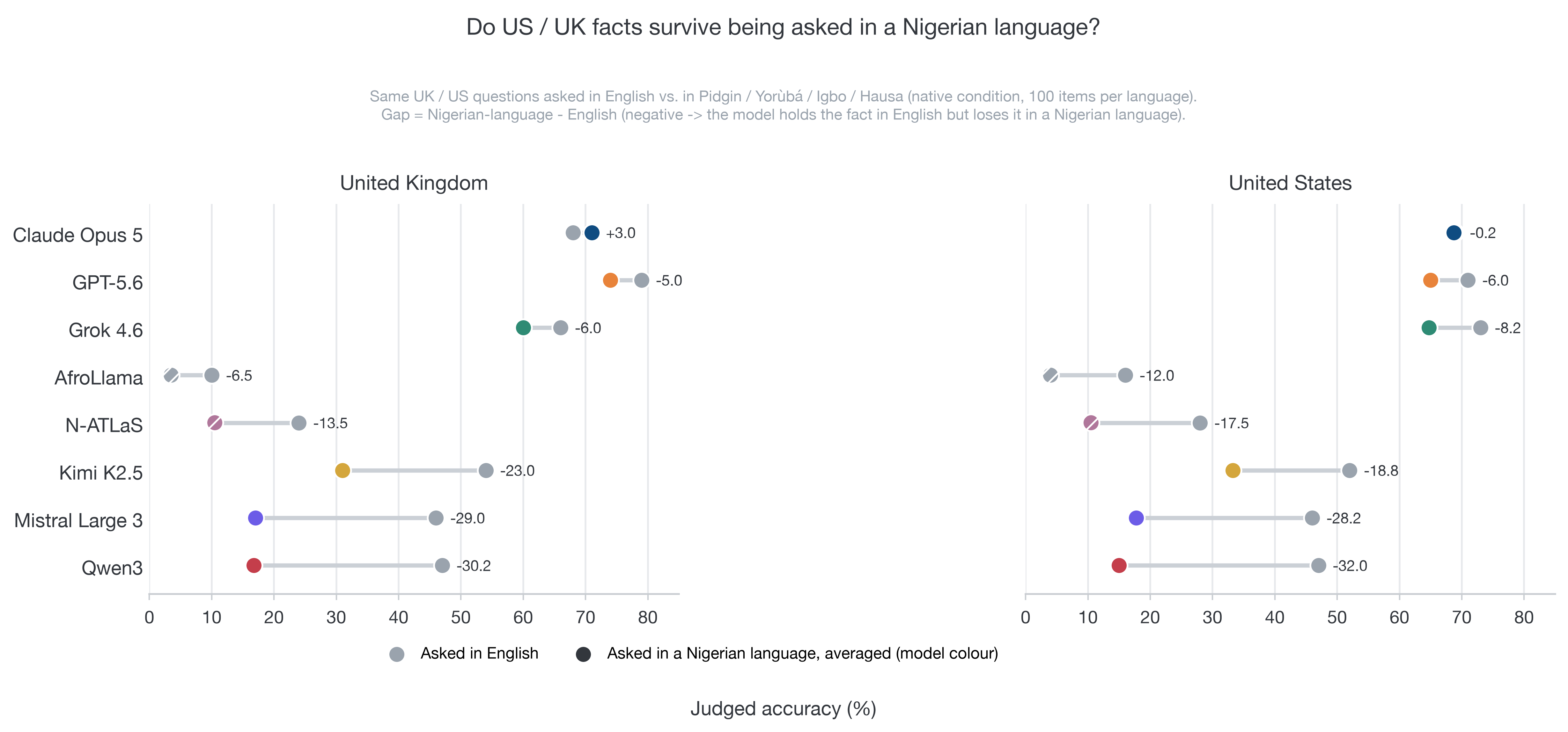}
  \caption{What a Nigerian query language costs on facts the models demonstrably
  hold. For each model, accuracy on the UK (left) and US (right) sets when the
  question is asked in English (grey) versus averaged over the four Nigerian
  languages (colour). Unlike Figure~\ref{fig:recovery}, where the gaps vanish
  because the knowledge is missing in the first place, here the gaps are large
  and negative for the weaker models, and they widen as overall capability
  falls: Claude Opus 5 and GPT-5.6 carry foreign facts across languages nearly
  intact ($-6$ to $+3$ points), whereas Qwen3 and Mistral Large 3 lose roughly
  two thirds of theirs ($-28$ to $-32$). This is the language effect measured
  where there is knowledge to lose.}
  \label{fig:langcost}
\end{figure}

The figures below add the per-model detail the body figures average away.
Figure~\ref{fig:bars} gives the language effect model by model,
Figure~\ref{fig:recovery} the answer-in-English comparison of
Section~\ref{sec:eng-answer}, and Figure~\ref{fig:langcost} the same language
effect measured on the UK and US sets, where the models demonstrably hold the
knowledge. Per-item model outputs and judge verdicts for every run are released
with the code and data.

\newcommand{\insertcountryaccuracy}{%
\begin{table}[t]
\centering
\footnotesize
\begin{tabular*}{\textwidth}{@{\extracolsep{\fill}}l r r l r r l r r@{}}
\toprule
Country & $n$ & Acc. & Country & $n$ & Acc. & Country & $n$ & Acc. \\
\midrule
Algeria & 125 & 44.8 & Eswatini & 124 & 33.9 & Niger & 123 & 42.3 \\
Angola & 124 & 34.7 & Ethiopia & 125 & 38.4 & Nigeria & 125 & 36.0 \\
Benin & 125 & 44.8 & Gabon & 125 & 43.2 & Rwanda & 125 & 41.6 \\
Botswana & 125 & 33.6 & Ghana & 125 & 45.6 & Senegal & 125 & 36.0 \\
Burkina Faso & 123 & 39.8 & Guinea & 125 & 52.8 & Seychelles & 119 & 29.4 \\
Burundi & 125 & 45.6 & Guinea-Bissau & 122 & 33.6 & Sierra Leone & 125 & 45.6 \\
Cabo Verde & 123 & 32.5 & Kenya & 125 & 51.2 & Somalia & 125 & 41.6 \\
Cameroon & 125 & 42.4 & Lesotho & 125 & 39.2 & South Africa & 124 & 46.8 \\
Central African Rep. & 119 & 31.1 & Liberia & 125 & 40.0 & South Sudan & 125 & 34.4 \\
Chad & 118 & 28.8 & Libya & 120 & 30.8 & Sudan & 122 & 34.4 \\
Comoros & 123 & 37.4 & Madagascar & 125 & 33.6 & São Tomé \& Príncipe & 125 & 36.8 \\
Congo (Rep.) & 125 & 37.6 & Malawi & 125 & 39.2 & Tanzania & 125 & 48.0 \\
Côte d'Ivoire & 125 & 41.6 & Mali & 125 & 36.8 & The Gambia & 122 & 45.9 \\
Djibouti & 124 & 37.1 & Mauritania & 125 & 31.2 & Togo & 121 & 41.3 \\
DR Congo & 125 & 31.2 & Mauritius & 125 & 42.4 & Tunisia & 124 & 41.1 \\
Egypt & 125 & 47.2 & Morocco & 125 & 58.4 & Uganda & 124 & 54.0 \\
Equatorial Guinea & 125 & 39.2 & Mozambique & 124 & 39.5 & Zambia & 125 & 38.4 \\
Eritrea & 119 & 37.8 & Namibia & 122 & 42.6 & Zimbabwe & 125 & 48.8 \\
\bottomrule
\end{tabular*}
\caption{Per-country factual accuracy of Claude Opus 5 on the \naija{} cross-country set, judged by an LLM grader and listed alphabetically. $n$ is the number of verified items for that country, which ranges from 118 to 125 because ambiguous items were dropped during verification, so accuracy is reported as a proportion. Overall: 2681/6689 = 40.1\%.}
\label{tab:country-accuracy}
\end{table}%
}

\insertcountryaccuracy

\end{document}